\documentclass[runningheads]{llncs}
\usepackage{graphicx}

\usepackage{latexsym}
\usepackage{amssymb}
\usepackage{amsmath}
\usepackage{mathrsfs}
\usepackage{txfonts}

\usepackage{subcaption}

\usepackage{algpseudocode}
\usepackage[]{algorithm2e}

\begin{document}
\title{Explainable AI through the Learning of Arguments}
\author{Jonas Bei \and
David Pomerenke \and
Lukas Schreiner \and
Sepideh Sharbaf \and
Pieter Collins \and
Nico Roos$^1$}\footnotetext[1]{The authors thank Julien Havel for his contribution to the initial phase of the reported research.}

\authorrunning{J. Bei et al.}
\institute{Data Science and Knowledge Engineering, Maastricht University, \\ Maastricht, The Netherlands
\email{\{jy.bei,d.pomerenke,lj.schreiner,s.sharbaf\}@student.maastrichtuniversity.nl, \{pieter.collins,roos\}@maastrichtuniversity.nl}\\
\url{http://www.maastrichtuniversity.nl/dke/}
}
\maketitle              %
\begin{abstract}

\emph{Learning arguments} is highly relevant to the field of explainable artificial intelligence. It is a family of symbolic machine learning techniques that is particularly human-interpretable. These techniques learn a set of arguments as an intermediate representation. Arguments are small rules with exceptions that can be chained to larger arguments for making predictions or decisions.

We investigate the learning of arguments, specifically the learning of arguments from a `case model' proposed by Verheij \cite{verheijProofProbabilities2017}. The case model in Verheij's approach are cases or scenarios in a legal setting. The number of cases in a case model are relatively low. Here, we investigate whether Verheij's approach can be used for learning arguments from other types of data sets with a much larger number of instances. We compare the learning of arguments from a case model with the HeRO algorithm \cite{johnstonAlgorithmInductionDefeasible2003} and learning a decision tree.

\keywords{Explainable AI \and Argumentation \and Learning Arguments \and Data Mining}
\end{abstract}
\setcounter{footnote}{1}
\section{Introduction}\label{Introduction}

\paragraph{Explainable AI}

Artificial intelligence, in a societal context, is confronted with a variety of requirements that are recently being investigated by the research fields around explainable, responsible and socially aware artificial intelligence \cite{specialinterestgrouponartificialintelligenceDutchArtificialIntelligence2018}. Here, we are concerned with explainability, that is, making the criteria transparent that underlie the decision of an algorithm.

Explainability is also increasingly becoming a \textit{legal} requirement of algorithms. In many countries, which as of recently includes the Netherlands 
\cite{raadvanstateECLINLRVS2017,rechtbankdenhaagECLINLRBDHA2020}, administrative and judicative decisions that have been supported by an algorithm are required to be comprehensible for judges and citizens \cite{doshi-velezAccountabilityAILaw2019}. The General Data Protection Regulation of the EU (see \cite{goodmanEuropeanUnionRegulations2017}), as well as similar legislation in the United States gives citizens a right to explainability also towards companies; albeit only when important decisions such as credit status are involved.

Surveys have been undertaken as to which machine learning techniques are suitable for explainable artificial intelligence, according to a range of sub-criteria. The result is that decision trees and approaches based on deductive logic are the most suitable techniques \cite{arrietaExplainableArtificialIntelligence2020,waltlExplainableArtificialIntelligence2018}. Here we investigate the learning of arguments, which can be classified broadly as a deductive logic approach.

\paragraph{Benefits of arguments}

Arguments provide reasons for believing conclusions given data \cite{Toulmin58}. 
Providing arguments for conclusions, considering exceptions to these arguments, and putting multiple small arguments together to build larger, convincing arguments, is the human way of justifying things. Learning of arguments from data sets and using these arguments for future decision making will provide more transparency than black box approaches. This transparency is important in domains such as law, public administration, health care, etc, as well as to the discovery of scientific explanations. 

Decision making based on learned argument addresses three problems: \\
The first problem is the mentioned requirement of the explainability of the decision of the algorithm. An algorithm that substantiates its claims with arguments can, if the arguments are properly presented, be understood by a human. Thus, humans can detect potential errors in the algorithm’s decision, or, hopefully, verify that no such errors have been made. This increases trust between human and machine \cite{doshi-velezAccountabilityAILaw2019}.

The second problem is that experts (or even non-expert humans) may possess
some relevant knowledge that can improve learning form training data, such as known causal relationships between some of the attributes of the data. Machine learning systems that produce arguments can incorporate the knowledge of both the data set and the expert.

The third problem is that humans may pose certain requirements towards the justification of a decision that are in conflict with the training data. Important example are racial, sexual, and other biases, that may be present in the training data, and would lead to the perpetuation of discrimination (and hence, further biased data sets) in the future. In order to avoid vicious circles of discrimination, humans may wish to reject discriminatory decisions implied by the data. This may also be realized by discarding, for example, racially motivated arguments.

\paragraph{Research aims}

Verheij \cite{verheijProofProbabilities2017} proposed an approach for learning arguments from a 'case model'. The case model in Verheij's approach are cases or scenarios in a legal setting, and the number of cases in a case model are relatively low. 
We investigate whether Verheij's approach can be used for learning arguments from other types of data sets with a much larger number of instances. We compare the learning of arguments from a case model with another approach for learning arguments, the HeRO algorithm \cite{johnstonAlgorithmInductionDefeasible2003}, and with learning a decision tree.

\paragraph{Paper outline}

The next section describes the related work. Section \ref{sec:preliminaries} describes the preliminaries and  Section \ref{Methodology} describes our implementation of Verheij's approach \cite{verheijProofProbabilities2017} as well as the other approaches that we implemented for comparison. Section \ref{sec:experiments} describes experimental evaluation and Section \ref{sec:conclusion} concludes the paper.

\section{Related work} \label{related_work}

Here we give a concise overview on the most relevant related work. 

\paragraph{Argumentation}

The modern view of argumentation was introduced by Toulmin \cite{Toulmin58}. He describes an argument as a (defeasible) warrant for a claim / conclusion given some data / premises. One of the first argumentation systems based on this idea was developed by Pollock \cite{Pollock87}, who extended predicate logic with defeasible and undercutting rules. 
An important issue in argumentation systems that make use of defeasible information, is determining which arguments are valid. Dung \cite{dungAcceptabilityArgumentsIts1995} showed
that this problem can be described by an \emph{argumentation framework}, which is a couple consisting of a set of atomic arguments with an attack relation over the arguments. He defines three argumentation semantics for determining the set of valid arguments given an argumentation framework, namely, the grounded, stable and preferred semantics. 
Arguments learned by Verheij's approach can be evaluated using the grounded semantics, while arguments learned by the HeRo algorithm may require the preferred semantics.

\paragraph{Learning arguments}

Kakas and Michael \cite{kakasAbductionArgumentationExplainable2020} give an insightful overview on argumentation in machine learning, enumerating multiple use cases of arguments. Here, we are concerned with argumentation as the target language for learning. Within this use case, they distinguish two paradigms. In the first paradigm, arguments are potentially large monolithic rules that directly map input facts to output facts. This paradigm comprises decision lists, exception lists, inductive logic programming with exceptions, and random forest methods. In the second paradigm, arguments consist of multiple chained smaller arguments, with intermediate concepts connecting the arguments. The smaller arguments describe local relations, that is, relations that only involve a small number of attributes. Within this paradigm fall the \textit{NERD} algorithm \cite{michaelCognitiveReasoningLearning2016}, \textit{machine coaching}, and \textit{SLAP}.

Two algorithms are explicitly concerned with the mining of defeasible rules: Firstly, the \textit{DefGen} algorithm uses association rule mining, for which highly optimized algorithms for big data exist, and post-processes the output by applying relevance criteria \cite{governatoriApplicationAssociationRules2001}. This high-level structure can also be found in our \textit{Pruned Search} algorithm introduced in Section \ref{sec:pruned-search-meth}. Secondly, the \textit{HeRO} algorithm iteratively applies the criterion of \textit{information gain}, taking inspiration from decision list mining and covering rule algorithms. We have implemented the HeRO algorithm; see Subsection \ref{hero-meth}.

\paragraph{Other rule-based learning approaches}

Competing approaches for the explainable learning of rules are decision trees, relational learning and inductive logic programming, and probabilistic and causal networks.

While decision trees are equivalent to sets of classification rules \cite[ch. 3.4]{wittenDataMiningPractical2017}, \cite[p. 358]{hanDataMiningConcepts2011}, the rules to which they correspond are long and unstructured. Domain experts prefer to work with well-structured sets of arguments, which then can be easily transformed into decision trees for classification \cite{breidenbachTextCode2021}. The advantage of decision trees is their suitability for big data. Some of the mentioned disadvantages can be overcome by pruning the decision tree (see also Section \ref{dectrees-meth}).

Relational learning and inductive logic programming are concerned with the learning of first-order logic and logic program representations, respectively, which can potentially be downgraded to work on propositional logic or attribute-value representations \cite{deraedtLogicalRelationalLearning2008}. Usually, algorithms in these fields produce monotonic rules. These do also allow for the construction of arguments, but these arguments cannot defeat each other and are therefore less similar to everyday argumentation than arguments from defeasible rules. One possibility for simulating exceptions is to use an exception predicate for each rule that has an exception. \cite{dimopoulosLearningNonmonotonicLogic1995} explores the theory of non-monotonic logic programming, \textit{XHAIL} \cite{rayInferringProcessModels2007} and \textit{TAL} \cite{corapiInductiveLogicProgramming2010} provide algorithms.

Probabilistic networks are most suitable for reasoning with uncertainty. Causal networks present an improvement over probabilistic networks (and all other methods) by taking into account the causal relationships between the variables. 
Causal networks also allow for counterfactual reasoning \cite[ch. 13.5.2]{russellArtificialIntelligenceModern2020}. Moreover, experiments indicate that it is easier to reason causally than it is to reason diagnostically \cite[p.121-128]{kahnemanJudgmentUncertaintyHeuristics1982}. 

\paragraph{Propositionalization}

The representation of both data and hypotheses in Verheij's approach is restricted to propositional logic \cite{verheijProofProbabilities2017}. In this project we investigate an extension to input data with an attribute-value representation \cite{deraedtLogicalRelationalLearning2008}, including categorical and continuous attributes. Our approach here is to preprocess the input data by transforming continuous and categorical attributes into propositions. Some techniques for propositionalization are described in \cite{deraedtLogicalRelationalLearning2008}. The propositionalization techniques explored in this project are Equal-Width Binning, Equal-Depth Binning, K-Means and DBSCAN, where each of the algorithms has its respective strengths and weaknesses. Equal-Width Binning and Equal-Depth Binning are the approaches with the least complexity, and K-Means and DBSCAN are more complex.

\section{Preliminaries} \label{sec:preliminaries}

Here, we present Verheij's approach \cite{verheijProofProbabilities2017}, which uses the notion of a \textit{case model} and three different notions of arguments.

\paragraph{Case models}

A case model is a description of different scenarios or situations (the cases) that can occur in the world, together with a preferences ordering over the cases denoting their relative likelihood. Each case is distinguished by the propositions that follow from it. We can alternatively define a case as the most general proposition which entails the propositions that follow from the case. 

In this paper, a \emph{case} will be a set of literals (or equivalently, a conjunction of literals). In this way, we arrive at a Boolean (propositional) representation that is suitable for machine learning. We do this by interpreting a case as a data point in the training data.

\emph{Presumption of innocence} is an example of a case model from \cite{verheijProofProbabilities2017}. 
This case model has two cases, $\{\mathit{innocent, \neg guilty}\}$ and $\{\mathit{\neg innocent, guilty, evidence}\}$, where the first case is most preferred, that is, the first case has a higher probability.

\paragraph{Arguments}

An \emph{argument} is a couple $(P,C)$ consisting of a \emph{premise} $P$ and a \emph{conclusion} $C$, each of which is a set of literals (or equivalently, a conjunction of literals). Note that an argument need not be valid. Verheij \cite{verheijProofProbabilities2017} defines the three types of arguments: \emph{coherent} arguments, \emph{presumptively valid} arguments and \emph{conclusive} arguments. There holds a superset relation between the three types of arguments.

An argument is \emph{coherent} for a case model if there exists a case in which both the premise and the conclusion are true. Note that the premise can be an empty set of literals.
Examples of coherent arguments are: $(\emptyset,\{guilty\})$, $(\{evidence\},\{\neg innocent\})$, etc.

A coherent argument is a \emph{presumptively valid} argument if the conclusion is true in the most preferred case in which the premise is true, given the preference ordering over the cases. Note that the conclusion need not be true in less preferred cases in which the premise is true. Presumptively valid arguments are most interesting in the context of this project, since they can have exceptions and are thus very much like human arguments, which is desirable from an explainable AI perspective. We use the notation $P \rightsquigarrow C$ to denote a presumptively valid argument with premise $P$ and conclusion $C$. If the premise $P$ is an empty set of literals, the conclusion $C$ holds by default: $\rightsquigarrow C$.
Examples of presumptively arguments are: $\rightsquigarrow \{\neg guilty\}$, $\rightsquigarrow \{innocent\}$, $\{evidence\}\rightsquigarrow\{\neg innocent\}$, $\{innocent\} \rightsquigarrow \{\neg guilty\}$, etc.

An argument is \textit{conclusive} if the conclusion is true in every case where the premise is true. Clearly, conclusive arguments are also presumptively valid. Conclusive argument need not be conclusive in the sense of everyday language because there is no formal requirement on a case model that it describes all possible cases. We use the notation $P \to C$ to denote a conclusive arguments. 
Examples of conclusive arguments are:  $\{innocent\} \to \{\neg guilty\}$, $\{guilty\}\to\{\neg innocent\}$, etc.

\section{Learning of Arguments}\label{Methodology}

This section discusses the learning of arguments, specifically from data sets that specify possibly continuous values for attributes. We assume a set of attributes for which each instance of the data sets specifies the attribute values. 

\subsection{Discretization Techniques}\label{Section_Discretization_Techniques}

With the exception of decision trees, the rule-mining algorithms in this project cannot be trained on continuous data. Therefore, in order to apply the rule-mining algorithms to data sets, we must rely on data discretization techniques to preprocess the data before mining the rules. 

\paragraph*{Equal-Width Binning}
This algorithm is a comparatively simple binning technique. Here, the range spanned by the smallest and largest value of a feature (referred to as $min$ and $max$ respectively,) is divided into a number of bins $k$, where each of these bins have size $\frac{max-min}{k}$. To discretize, values are assigned to the respective bin they fall into.

\paragraph{Equal-Depth Binning}
Equal-depth or equal-frequency binning is another simple discretization approach. Here, values are assigned to one of $k$ bins, such that each bin approximately holds the same number of instances. This is done by sorting the values of the feature and assigning $\frac{n}{k}$ of the sorted instances into each bin, where $n$ is the number of total values.

\paragraph{Clustering approaches}

To discretize more complex features in the data, clustering approaches are considered. Here, values of a given feature in the data are clustered, and replaced by the discretized value. In the data set, clusters are represented as ranges, where each cluster is described by its smallest and largest value. By the nature of the given clustering algorithms, these ranges do not overlap.

\paragraph{K-Means Clustering}
K-Means \cite{LeastSquareLloyd} is based on the idea of centroids, which are points in the centre of the cluster. Here, $k$ centroids are initialized randomly, and the instances are assigned to the cluster whose centroid is closest. Then, the centroids are moved to the mean of the cluster, and the instances are assigned to their new cluster. The algorithm converges when the movement of centroids is below a certain threshold.

\paragraph{DBSCAN Clustering}

DBSCAN  by \cite{DBSCANPaper} considers clusters to be regions of high density. For each instance, the algorithm counts the number of instances within a distance $\epsilon$, also called the instance's $\epsilon$-neighbourhood. If this number of neighbours of an instance surpasses a given threshold, the instance is considered to be a core instance, an instance within a dense region. The neighbours of this core instance are considered to be in the same cluster, where some neighbours may also be core instances themselves. Therefore, a cluster consists of a multitude of core instances.

\paragraph{Cluster Optimization}

The aforementioned clustering algorithms all provide parameters that can be tuned in order to find  clusters representing the data correctly. In this project, the silhouette score introduced by \cite{ROUSSEEUW198753} has been utilized to provide a metric for accuracy of clusters. This score computes the mean silhouette coefficient of all samples: $silhouette\_ score = \frac{b - a}{\max(a, b)}$. Here, $a$ denotes the mean distance to the other instances in the same cluster (intra-cluster distance) and $b$ denotes the minimal distance to another instance that is not part of the same cluster (nearest-cluster distance).

Clusters are optimized by exhaustive search in this project, i.e., every combination of parameters is tested using the silhouette score, before returning the parameters resulting in the highest score.

\subsection{Algorithms for learning arguments}\label{Section_Learning_Arguments}

We have implemented three different algorithms for learning arguments from data.\footnote{Our code is available as an open source Python module at: \\ https://github.com/learning-arguments/
learning\_arguments} The first  algorithm is devised by ourselves, the second one is implemented by ourselves according to the high-level description in \cite{johnstonInductionDefeasibleLogic2003}, and the third one is based on the open-source library \textit{scikit-learn} \cite{pedregosa2011scikit}.

\subsubsection{Pruned Search} \label{sec:pruned-search-meth}

\begin{figure}[htb]
	\centering
	\includegraphics[width=0.6\textwidth]{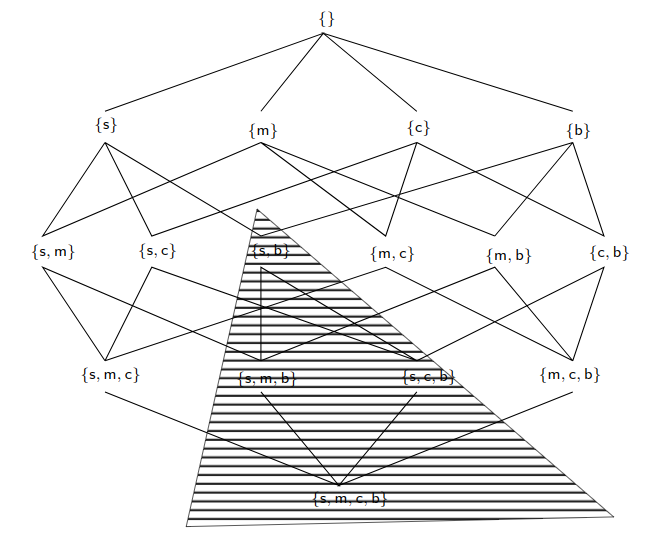}
	\caption{\textit{Pruning specializations.} From \cite[p.~52]{deraedtLogicalRelationalLearning2008}. All $2^n$ subsets of $\{s, m, c, b\}$ are systematically searched, starting from the most general set at the top. Knowing that the set $\{s, b\}$ is infrequent allows us to prune all its specializations, which is a lot. }
	\label{fig:pruning}
\end{figure}

\noindent A naive implementation of Verheij's approach is not very efficient and has a worse case time complexity of $n^k$ where $k$ is the number of attributes of the data set and $n$ is the number of bins. 
The \emph{Pruned Search} algorithm improves the run-time by pruning the search space in a systematic way. This technique is known from frequent pattern mining (and its application to association rule mining \cite{agrawalFastAlgorithmsMining1994}), and is described in the context of logical learning in \cite{deraedtLogicalRelationalLearning2008}. The idea is to identify a quality criterion, for which the following is true: If a set fulfills the quality criterion, all its subsets must also fulfill the quality criterion. (Alternatively: If a set fulfills the quality criterion, all its \textit{super}sets must also fulfill the quality criterion; this can be visualized by "flipping" the search space or the direction of the search). For example, in the context of frequent pattern mining, if a set is frequent, then all its subsets must also be frequent. The principle of pruning the search space is visualized in Figure \ref{fig:pruning}.

This raises the issue of the selection of a suitable quality criterion for pruning the search space in our application of learning arguments. We found two quality criteria: 

\begin{enumerate}
	\item \textit{If an argument $(P,C)$ is conclusive, then all coherent arguments $(P', C)$ must also be conclusive, for all $P'$ that are a superset of $P$.} 
	\item \textit{If an argument $(P,C)$ is coherent, then all arguments $(P',C)$ must also be coherent, for all $P'$ that are a subset of $P$.}
\end{enumerate}

The most important part of the learning algorithm in terms of efficiency is the learning of presumptively valid arguments: They are relevant for a prediction, and there are usually many more  presumptively valid arguments than conclusive arguments.
Unfortunately, we can prove that being presumptively valid is not a quality criterion  that can be used to prune the search space. The underlying reason is that presumptively valid arguments can be overruled by more specific arguments.

Although we cannot use presumptive validity itself as a quality criterion for pruning the search space, we can at least use a condition for presumptive validity, namely: \emph{coherence}. Our algorithm starts a search for each literal (each combination of an attribute and a bin), looking for coherent arguments with this literal as a conclusion. We search for premises with increasing number of literals. After the search is completed, we filter and merge the resulting arguments.

\begin{enumerate}
	\item The filtering step is necessary for removing irrelevant rules. For example, when there are two arguments $a \rightsquigarrow d$ and $a \land b \land c \rightsquigarrow d$, then the second argument is more specific than the first argument and therefore only relevant if there is another relevant argument, such as $a \land b \rightsquigarrow \neg d$, to which it is an exception. Generally speaking, an argument A is relevant if there is no less specific argument to A, or if A is an exception to a relevant argument. We say that an argument $P_1 \rightsquigarrow c_1$ is more specific than another argument $P_2 \rightsquigarrow c_2$, if its premise $P_1$ are a proper superset of the premise $P_2$ of the other argument. 
	\item During the argument generation, we only generate arguments with a conclusion of a single literal. We reduce the number of arguments for legibility by merging together any arguments $(P_1, c_1)$, ..., ($P_m, c_m)$ that have the same premises $P_1=...=P_m$ to a single argument $(P_1= ...= P_2, \{c_1, ..., c_m\})$ with multiple conclusions.
\end{enumerate}

At the end of the search step we gather all coherent arguments of the step, and check all combinations of these arguments whether their premises differ in exactly two literals. The reason is: If they are different in two literals, we can take the union of the premises as a new premise of size $i+1$, and we know that many subsets of this premise lead to a coherent argument. Consider, for example, two premises $\{a,b,c,d\}$ and $\{b,c,d,e\}$: The union is $\{a,b,c,d,e\}$ with size $n=5$. Enumeration shows that $2^n-2^{n-2} = 24$ of its subsets are also subsets of at least one of the two premises of which we already know that they are coherent. It is thus much more likely for the new premise to also lead to a coherent argument than it would be for an arbitrary premise. We use this observation as a heuristic to speed up the search for other coherent arguments.

The principle of combining small sets fulfilling the quality criterion into larger sets likely to fulfill the quality criterion is known from the Apriori algorithm \cite{tanIntroductionDataMining2014}. It makes the Apriori algorithm suitable for big data sets. Here, because coherence is only a condition but not the same as presumptive validity (which we are looking for), it at least makes the algorithm efficient enough for the medium-sized data sets we use.

An argument that is presumptively valid but not conclusive will have exceptions. We recursively search for exceptions on each presumptively valid argument, as well as exception on exceptions on exceptions etc., till a maximum specified depth.

\subsubsection{HeRO algorithm} \label{hero-meth}

The HeRO algorithm has been devised by \cite{johnstonAlgorithmInductionDefeasible2003}, and the research behind it, like \cite{verheijProofProbabilities2017}, is also originally targeted towards the legal domain \cite{johnstonInductionDefeasibleLogic2003}. It does not primarily perform a systematic search, but rather an incremental search: At each step, it considers which argument would be most valuable to be added to the theory in order to increase the accuracy the most; and then it adds the most valuable argument to the theory and asks the question again, until there is no more argument that can increase the accuracy. 

The algorithm builds up a totally ordered set of arguments, and at every step it considers all positions (before, after, or between the existing arguments) for adding the next argument. For determining the most valuable argument (and its most valuable position), the criterion of \textit{information gain}, that is increase in accuracy on the training set, is used. Similar to the Pruned Search algorithm presented above, the HeRO algorithm also starts by considering simple arguments and then in some cases also considers arguments where the premise is more specific. The mechanism for deciding whether a more specific premise should also be considered uses the criterion of \textit{maximum information gain}. The maximum information gain of an argument is the highest information gain that can be achieved by any argument that is more specific. This is equivalent to the information gain that would be achieved by adding an argument that correctly predicted all the rows where the premises hold.

\subsubsection{Decision tree algorithm} \label{dectrees-meth}

To classify by building tree models, the open-source library \textit{scikit-learn} \cite{pedregosa2011scikit} is used. This implementation utilizes the CART \cite{DecTreePaper} algorithm. Here, the tree is built choosing a feature $k$ and a threshold $t_k$ by using a cost function measuring the purity of the subsets produced by the split. In this project, this is measured by the Gini impurity introduced by \cite{DecTreePaper}. Once the split has been made, the algorithm iteratively splits the subsets further, until a given maximum depth is reached, or no split reducing impurity can be found.

In a decision tree, the nodes at the bottom of the tree are referred to as leaf nodes. Trees can be converted into decision rules, where each leaf node is associated with one rule. Here, the path traversed through the decision tree represents the premise that must hold for the conclusion at the child node.

To maximize performance of the decision tree algorithm, various hyper-parameters can be tuned. In this project, this is done via Bayesian Optimization \cite{BayesOptPaper} utilizing the \textit{scikit-optimize} package \cite{skopt}. This algorithm samples points to construct an interpolation function, also called posterior function. This function represents the objective function (which, in this case, is a function measuring the accuracy of the tree with its parameters as inputs). New points are found using an acquisition function, which balances exploration and exploitation by calculating uncertainty in the posterior function. These query points are then used to update the posterior function. After a given number of iterations, the algorithm converges, returning an estimate of the optimal parameters by using the posterior function.

\section{Experimental evaluation} \label{sec:experiments}

\subsection{Experimental setup}

\subsubsection{Legal examples}

We have evaluated the Pruned Search and the HeRO algorithm on legal examples described in \cite{verheijProofProbabilities2017} and \cite{verheijAnalyzingSimonshavenCase2020}.

\subsubsection{Boston Housing Dataset}

We have evaluated all algorithms on the Boston Housing Dataset\footnote{http://lib.stat.cmu.edu/datasets/boston}. The Boston Housing Dataset specifies the values of 14 attributes for 506 instances. We evaluated the performance on this data set in combination with descretization algorithms. The main parameters were the number of bins used.
An optimization algorithm for finding the ideal number of bins has been implemented. Because the search algorithms are very sensitive to the number of bins, we also ran the discretization algorithms with  predefined number of bins, namely 2, and 4 bins.

Binning implies that several data points of the data set are grouped together. Assuming that all data points in the data set are equally likely, the number of data points that are grouped together determine the relative likelihood that we need for the case model.

The four main steps of the experiments are data preprocessing, model training, predictions, as well as model evaluation. In a first step, the data is discretized by a method described in section \ref{Section_Discretization_Techniques}. In the experiment using decision trees, only the target column is discretized. In a second step, the selected algorithm for learning the arguments from the data is applied. Afterwards, the learned model is used to generate predictions from the training data as well as the test data. Finally, the predictions are evaluated by computing accuracy and weighted F1-score. The training time is measured in order to get an understanding of the relative computational cost of the algorithms.

\paragraph{Parameter tuning}
The Pruned Search algorithm has two hyper-parameters that needed to be tuned. Next to the search depth for exception on exceptions etc., which is tested with the values 1, 5 and 20, the values 2 and 4 are tested for the maximum premise size constraint. A priori, we assume that the former will have a significant impact on the run-time while the latter will mainly determine the quality of the predictions. 

Although the decision tree algorithm optimizes the parameters by Bayesian Optimization, there is still a need for specifying the parameter search space. Here, the maximum number of features randomly chosen at a split can be set between 1 and the number of features of the training data. The maximum depth is capped at 50 to retain explainability and the minimum number of samples required at a leaf node is constrained between 1 and 1000. The minimum number of samples required to split an internal node is between 2 and 1000. 

The HeRO algorithm does not require any hyper-parameter tuning.

\subsection{Results}

\subsubsection{Legal examples}

The experimental results show that the Pruned Search algorithm finds all arguments mentioned in the papers \cite{verheijProofProbabilities2017} and \cite{verheijAnalyzingSimonshavenCase2020}. It also finds quite a few additional arguments that are correct but often irrelevant.

The HeRO algorithm generates a more concise set of arguments. However, the arguments can imply counter-intuitive self-attacks. Consider for instance the first case model in \cite{verheijProofProbabilities2017}: \emph{Presumption of innocence}. 
This case model has two cases, $\{\mathit{innocent, \neg guilty}\}$ and $\{\mathit{\neg innocent, guilty, evidence}\}$, where the first case is most preferred. 
HeRO determines the following two arguments: $\mathit{\rightsquigarrow innocent \wedge \neg guilty \wedge evidence}$ and \\ $\mathit{evidence \rightsquigarrow \neg innocent \wedge guilty}$, which imply a self-attacking argument. 
The first argument is counter-intuitive. HeRO determines this argument because if any information is given regarding whether or not there is evidence, then indeed it will be the information that there is evidence.
Note that self-attacking arguments imply that we need to use Dung's preferred semantics \cite{dungAcceptabilityArgumentsIts1995} for determining the set of valid arguments.

\subsubsection{Boston Housing Dataset}

We trained the three algorithms using 80\% of the Boston Housing Dataset. The remaining 20\% were used to test the models learned by the algorithms. We evaluated the algorithms on both the training and the test data set. 

\begin{table}
	\caption{Summary of the Decision Tree results. The `--' denotes that there is no dependence on the parameter.}
	\label{tab:dt}
	\centerline{
	\begin{tabular}{|l|l|c|c|c|c|}
		\hline
		data type & binning method & \# bins & search depth & accuracy & F1 \\
		\hline\hline
		training & kMeans & -- & -- & 0.8613 & 0.8634 \\
		test & kMeans & -- & -- & 0.8613 & 0.8634 \\
		\hline
		training & DBSCAN & -- & -- & 0.9975 & 0.9963 \\
		test & DBSCAN & -- & -- & 0.9975 & 0.9963 \\
		\hline
		training & other methods & -- & -- & 1 & 1 \\
		test & other methods & -- & -- & 1 & 1 \\
		\hline
		\end{tabular} }
\end{table}

\paragraph{Decision Trees} When using the Boston Housing Data set, the decision trees were scoring a perfect accuracy of 1 when using equal-depth binning or equal-width binning. Using DBSCAN gave a slightly lower accuracy of 0.99 and using k-means yielded 0.86; see Table \ref{tab:dt}. Those figures showcase very well the impact and importance of choosing a good technique when discretizing the data. Note that the accuracy alone does not provide a complete picture of the quality of the algorithm: For example, using one bin for all the data would result in an accuracy of 1, yet the algorithm would not explain any structure in the data. With regard to the training time, the decision trees run significantly longer than the Pruned Search algorithms. When only discretizing the target column using equal-width binning and leaving the input values continuous, the decision trees achieve an accuracy of 0.92. This indicates that the decision trees are able to capture the structure well. 

\begin{table}[t]
	\caption{Summary of the Pruned Search results. The `--' denotes that there is no dependence on the parameter. The optimized number of bins is denoted 'opt' in the table. Note that the optimal number of bins may be different for each column of the data set and may depend on the binning method.}
	\label{tab:ps}
	\centerline{
		\begin{tabular}{|l|l|c|c|c|c|c|}
			\hline
			data type & binning method & \# bins & search depth & max \# premises & accuracy & F1 \\
			\hline\hline
			training & kMeans & 2 & -- & -- & 0.7772 & 0.6798 \\
			test & kMeans & 2 & -- & -- & 0.8039 &0.7165 \\
			training & kMeans & opt & -- & -- & 0.8515 & 0.8537 \\
			test & kMeans & opt & -- & -- & 0.8168 & 0.8137 \\
			\hline
			training & DBSCAN & -- & -- & -- & 0.9530 & 0.9335 \\
			test & DBSCAN & -- & -- & -- & 0.9706 & 0.9561 \\
			\hline
			training & EWBinning & 2 & -- & -- & 0.9431 & 0.9154 \\
			test & EWBinning & 2 & -- & -- & 0.9118 & 0.8697 \\
			training & EWBinning & 4 & -- & -- & 0.9431 & 0.9154 \\
			test & EWBinning & 4 & -- & -- & 0.9118 & 0.8697 \\
			training & EWBinning & opt & -- & -- & 0.9431 & 0.9154 \\
			test & EWBinning & opt & -- & -- & 0.9118 & 0.8697 \\
			\hline
			training & EDBinning & 2 & -- & -- & 0.8317 & 0.8317 \\
			test & EDBinning & 2 & -- & -- & 0.8333 & 0.8334 \\
			training & EDBinning & 4 & -- & -- & 0.8243 & 0.8226 \\
			test & EDBinning & 4 & -- & -- & 0.8725 & 0.8132 \\
			training & EDBinning & opt & -- & -- & 0.8168 & 0.8137 \\
			test & EDBinning & opt & -- & -- & 0.7353 & 0.7318 \\
			\hline
	\end{tabular} }
\end{table}

\paragraph{Pruned Search} When it comes to Pruned Search, the results also exhibit high results for the accuracy and F1 scores. The average accuracy (F1 score) on the training set is 0.88 (0.83) and 0.85 (0.83) on the test set; see Table \ref{tab:ps}. We ran 198 experiments with the Pruned Search algorithm. The standard deviation of the evaluation metrics (accuracy: 0.0868, F1: 0.0864) indicate that the algorithms performance is rather robust. 

Table \ref{tab:correlation} shows the correlation between the hyperparameters as well as the accuracy and F1 score.  The results are based on the test set and training set together. When studying the correlations, we noticed that the Pruned Search algorithms do not significantly vary in accuracy and F1 score when adjusting search depth and maximum number of literals in a premise. This observation is contrary to our initial hypothesis. The positive correlation between the maximum number of literals in a premise and the run-time suggests that it increases the computational complexity. The number of bins are also positively correlated with the run-time, yet exhibit a negative correlation on the accuracy metrics. Since fewer bins make the problem easier for the algorithm, this does not come as a surprise. A very interesting observation is the negative correlation of the run-time and accuracy / F1 score. This indicates that simpler and faster algorithms perform better on this data set, likely because they are less predisposed to overfitting. 

As mentioned, the discretization algorithm has a significant impact on the success of the algorithm. When looking at the situation where Pruned Search is used to mine the arguments and the number of bins is fixed to 2, one can observe that the accuracy on the test set increases when using the discretization algorithms in the following order: k-means (average accuracy 0.80), equal depth binning (0.83), equal width binning (0.91), DBSCAN (0.97). The ordering is strict, meaning that using a different discretization algorithm will always yield a higher or lower accuracy in the given settings. This emphasizes the  significant impact of binning on the algorithm's performance. 

\begin{table}
	\caption{Summary of the HeRO results. The `--' denotes that there is no dependence on the parameter. The optimized number of bins is denoted 'opt' in the table. Note that the optimal number of bins may be different for each column of the data set and may depend on the binning method.}
	\label{tab:hero}
	\centerline{
		\begin{tabular}{|l|l|c|c|c|c|c|}
			\hline
			data type & binning method & \# bins & search depth & max \# premises & accuracy & F1 \\
			\hline\hline
			training & kMeans & -- & -- & -- & 0.8039 & 0.7165 \\
			test & kMeans & -- & -- & -- & 0.7772 & 0.6798 \\
			\hline
			training & DBSCAN & -- & -- & -- & 0.9706 & 0.9561 \\
			test & DBSCAN & -- & -- & -- & 0.9455 & 0.9191 \\
			\hline
			training & EWBinning & 2 & -- & -- & 0.9431 & 0.9154 \\
			test & EWBinning & 2 & -- & -- & 0.9118 & 0.8697 \\
			training & EWBinning & 4 & -- & -- & 0.8861 & 0.8326 \\
			test & EWBinning & 4 & -- & -- & 0.8725 & 0.8132 \\
			training & EWBinning & opt & -- & -- & 0.9431 & 0.9154 \\
			test & EWBinning & opt & -- & -- & 0.9118 & 0.8697 \\
			\hline
			training & EDBinning & 2 & -- & -- & 0.5392 & 0.3778 \\
			test & EDBinning & 2 & -- & -- & 0.5 & 0.3333 \\
			training & EDBinning & 4 & -- & -- & 0.5392 & 0.3778 \\
			test & EDBinning & 4 & -- & -- & 0.5 & 0.3333 \\
			training & EDBinning & opt & -- & -- & 0.5817 & 0.4278 \\
			test & EDBinning & opt & -- & -- & 0.5588 & 0.4007 \\
			\hline
	\end{tabular} }
\end{table}

\paragraph{HeRO} The HeRO algorithm behaves similar compared to the Pruned Search algorithm in terms of performance; see Table \ref{tab:hero}. Similar as outlined above, the discretization algorithm is the main driver for the algorithm's performance. While the equal-depth binning yields an average accuracy (F1) of only 0.53 over all experiments, using k-means improves the results already significantly with an average accuracy of 0.79 (0.70). Equal-width binning further improves the situation by yielding 0.91 (0.86) and with an average accuracy 0.95 (0.93), DBSCAN gives the best results for the HeRO algorithm.

\begin{table}[h]
	\caption{Pruned Search Hyper-parameter Correlation Table}
	\label{tab:correlation}
	\centering
	\begin{tabular}{|l|cccccc|}
		\hline
		\textit{n=198} & \textit{Acc} & \textit{F1} & \textit{\# bins} & \textit{Depth} & \textit{Run-time} & \textit{Max \# prem.}                                   \\ \hline\hline
		Acc                                                                               & 1.000        &             &                   &                &                  &                                                 \\
		F1                                                                                & 0.940        & 1.000       &                   &                &                  &                                                 \\
		\# bins                                                                          & -0.008       & 0.057       & 1.000             &                &                  &                                                 \\
		Depth                                                                             & 0.000        & 0.000       & 0.000             & 1.000          &                  &                                                 \\
		Run-time                                                                           & -0.173       & -0.001      & 0.170             & 0.035          & 1.000            &                                                  \\
		Max \# prem.                                                                      & 0.000        & 0.000       &                   &                & 0.207            & 1.000                                           \\ \hline
	\end{tabular}
\end{table}

\subsection{Discussion}

The experiments show that arguments learned from a case model enables accurate predictions, yet needs further efforts to become practically applicable. There are two main issues that the experimental results bring to light. The first one is the exponentially increasing computational complexity of both the search and discretization algorithms. These limitations should be  addressed first.

Another point worth mentioning is the binning itself. In cases where the data is binned in very few bins, it can happen that the data is heavily skewed due to outliers. When e.g. 95\% of the houses are categorized as 'high price', the algorithm will score a very high accuracy with a naive prediction of always predicting 'high price'. It is obvious that the ability of the algorithms to explain patterns in data will decrease if the number of bins is reduced, while accuracy tends to increase. For that reason, just considering the accuracy might lead to false conclusions.

Furthermore, the experiments showed that simpler algorithms seem to do better than the more complex algorithms. The key takeaway from this may be that learning arguments tends to over-fit quickly.

\section{Conclusion} \label{sec:conclusion}

We have implemented Verheij's approach \cite{verheijProofProbabilities2017} for learning arguments from a case model and showed that (1) it can reproduce the examples given in \cite{verheijProofProbabilities2017} and \cite{verheijAnalyzingSimonshavenCase2020}, and (2) it can also be used to learn arguments from a data set consisting of instances specifying values of attributes. However, the implementation of Verheij's approach produces many correct but irrelevant arguments. A serious limitation is the run time of our implementation. To make the approach applicable to larger data sets, further research in reducing the run time is needed. Finally, the accuracy of the learned arguments for the Boston Housing Dataset depends on the used discretization algorithm with DBSCAN giving the highest performance.

We also implemented the HeRO algorithm \cite{johnstonAlgorithmInductionDefeasible2003} for comparison. The HeRO algorithm does not learn irrelevant arguments because it is employing the criterion of information gain. However, the learned arguments are not always intuitively plausible and may imply self attacking arguments. The accuracy of the learned arguments for the Boston Housing Dataset is 4\% less compared to the implementation of Verheij's approach. Moreover, HeRO is more sensitive w.r.t.\ the choice of the discretization algorithm, with DBSCAN giving the best performance. 

The decision tree algorithm that we implemented uses pruning on the learned tree to discard less relevant nodes. The arguments implied by the decision tree are not very intuitive. However, the decision tree algorithm reaches an accuracy of 100\%.

 \bibliographystyle{splncs04}
 \bibliography{bibliography}

\end{document}